\pdfoutput=1
\documentclass[11pt,a4paper]{article}
\usepackage[hyperref]{emnlp2020}
\usepackage{times}
\usepackage{latexsym}
\usepackage{adjustbox}
\usepackage{amsmath,amssymb}

\usepackage{microtype}

\aclfinalcopy 

\title{CUT: Controllable Unsupervised Text Simplification}

\author{Oleg Kariuk \\
  Ukrainian Catholic University \\
  ShipHawk \\
  \texttt{olegkariuk@gmail.com} \\\And
  Dima Karamshuk \\
  Facebook \\
  \texttt{karamshuk@fb.com} \\
}
\date{}

\begin{document}

\maketitle

\begin{abstract}

In this paper, we focus on the challenge of learning controllable text simplifications in unsupervised settings. While this problem has been previously discussed for supervised learning algorithms, the literature on the analogies in unsupervised methods is scarse. We propose two unsupervised mechanisms for controlling the output complexity of the generated texts, namely, \emph{back translation with control tokens} (a learning-based approach) and \emph{simplicity-aware beam search} (decoding-based approach). We show that by nudging a back-translation algorithm to understand the relative simplicity of a text in comparison to its noisy translation, the algorithm self-supervises itself to produce the output of the desired complexity. This approach achieves competitive performance on well-established benchmarks: SARI score of $46.88\%$ and FKGL of $3.65\%$ on the Newsela dataset.

\end{abstract}

\section{Introduction}

Text simplification deals with the problem of rewriting complex texts into a language which is easier to read and understand while preserving its original information and meaning. Simplification techniques can improve reading comprehension for a broader range of users, ranging from foreign language learners \cite{allen-role-of-relative-clauses, Petersen2007TextSF} and non-experts \cite{Elhadad:2007:MLT:1572392.1572402, Siddharthan:2010:RDC:1857999.1858142} to people with disabilities \cite{Canning:2000:CGS:647238.720905, carroll-etal-1999-simplifying} or low-literacy \cite{de-belder-text-simplification-for-children}. 

A variety of supervised and unsupervised approaches have been recently applied to the text simplification problem. On the supervised side, it has been considered as a monolingual machine translation exercise where a number of dedicated  \emph{sequence-to-sequence} models have been proposed \cite{kajiwara-komachi-2016-building, scarton-etal-2018-text, zhang-lapata-2017-sentence}. Unfortunately, the scarcity of parallel datasets limits the scalability of these approaches in application to different languages, domains, and output styles. Moreover, the \emph{Parallel Wikipedia Simplification} corpus, which has become the benchmark dataset for training and evaluating text simplification systems, is (a) prone to automatic sentence alignment errors, (b) contains a lot of inadequate simplifications and (c) poorly generalizes to other text styles \cite{xu-etal-2015-problems}.  

Inspired by the recent success \cite{lample2017unsupervised} of unsupervised machine translation, \cite{zhao2020semi} have applied unsupervised techniques of \emph{back-translation} and \emph{noisy auto-encoders} to the problem of text simplification. By adjusting these mechanisms to the particularities of mono-lingual translation, they managed to achieve performance results competitive with the supervised models. While this surprising effectiveness of unsupervised approaches and their ability to learn from vastly available non-parallel text corpora make them a significantly more attractive option than supervised text simplification, there yet remains many unsolved challenges. One of such is the challenge of \emph{controlling the complexity} of the output produced by a text simplification algorithm. This problem has been solved for supervised approaches by learning on the characteristics of the output texts \cite{scarton2018learning, martin2019controllable}, an idea that does not immediately apply to unsupervised settings. 

In this work, we build on an idea of using dedicated tokens \cite{martin2019controllable} to control the complexity of the produced output texts and apply it to unsupervised text simplification. We introduce this mechanism to the back translation algorithm, which allows the model to self-supervise the process of learning inter-relations between a control sequence and the complexity of the produced output. We compare this technique with \emph{simplicity-aware penalties} for beam-search generation of the output texts, thus leveraging on both \emph{learning-based} and \emph{decoding-based} mechanisms for controlled text generation \cite{kikuchi2016controlling}. Together these contributions allow us to achieve and exceed the previously reported SARI and FKGL results on the Newsela dataset. 

\section{Methodology}

We propose two different approaches for controlling the output complexity of unsupervised text simplification algorithms, namely, \emph{back-translation with control tokens} (a learning-based approach) and \emph{beam search with simplicity-aware penalties} (a decoding-based approach).  

\subsection{Back-translation with control tokens}

For learning-based control, we get inspiration from \cite{martin2019controllable} who introduced different types of dedicated tokens to control the output complexity of the supervised text simplification. We apply this idea to a self-supervised back-translation algorithm~\cite{sennrich2016improving} as follows. 

Firstly, we produce the noisy translation $u^*(y)$ of the original phrase $y$ and compute control tokens based on the original text $y$ and its noisy translation, i.e., $H(y, u^*(y))$, where $H$ is a sequence of four tokens which represent the compression ratio, Levenshtein similarity, word rank and the depth of dependency tree as defined in \cite{martin2019controllable}.  

Then we concatenate the computed control tokens with the noisy translation ${H(y, u^*(y)) \frown u^*(y)}$ and translate the resulting sequence back, aiming at reproducing the original text. We apply this procedure for both -- simple ($y \sim S$) and complex sentences ($x \sim C$) -- and train the algorithm to minimize the difference between the original and the output texts, i.e., 

\begin{equation*}
    \begin{split}
\mathcal{L} = 
& \mathbb{E}_{y \sim S}[-\log P_{c \rightarrow s}(y|H(y, u^*(y)) \frown u^*(y))] + \\
                     & \mathbb{E}_{x \sim C}[-\log P_{s \rightarrow c}(x|H(x, v^*(x)) \frown v^*(x))]        
    \end{split}
\end{equation*}


\subsection{Simplicity-aware beam search}

For a decoding-based control, we introduce the simplicit-aware penalties in the beam search when generating the output texts. Instead of decoding the most probable words in a greedy fashion, the beam search algorithm generates an output sentence by keeping a fixed number (specified by beam size parameter) of hypotheses with the highest log-probability at each step. 

To manage the exact matches ratio, length and simplicity (FKGL-based) of each hypothesis in the beam search iterations, we added three types of score penalties: \newline

\textbf{Length penalty} (LP) favors shorter or longer hypothesis depending on $\lambda_{length}$ parameter:
\[LP = e^{\lambda_{length} \times length(hypothesis)}\]

\textbf{Exact matches penalty} (EMP) uses cosine similarity between input and hypothesis to restrict the copying of input:
\[EMP = e^{\lambda_{exact\_matches} \times cos(input, hypothesis)}\]

\textbf{FKGL penalty} (FKGLP) encourages hypothesis with lower FKGL score:
\[FKGLP = e^{\lambda_{FKGL} \times  FKGL(hypothesis)}\]

We find an optimal combination of different penalties by running a grid search on already trained models over a hold-out validation dataset and optimizing for the best trade-off between SARI and FKGL~\footnote{LP=0.1, EMP=0.4, LFGKLP=0.4 and LP=0.4, EMP=1.3, LFGKLP=1.0 have been identified as optimal for CUT-S (p) and CUT-U (p) in $range(0.1, 1.3, 0.3)$, correspondingly.}. 

\subsection{Pre-trained language models}

We use the pre-trained language models from XLM library \cite{lample2019cross} as the basis for our experiments. The XLM models were trained on a large Wikipedia + Toronto Book Corpus and already encapsulate powerful embeddings for English texts. We also use the implementation of the \emph{noisy auto-encoders} (AE) and \emph{back-translation} (BT) from the XLM library as the basis for our unsupervised approach and use supervised \emph{machine translation} (MT) step for comparing the performance of the supervised and semi-supervised settings. We denote vanilla unsupervised and semi-supervised XLM models with \emph{XLM-U} and \emph{XLM-S}; our approach to back-translation with control tokens with \emph{CUT-U (t)} and \emph{CUT-S (t)} and a variant with simplicity-aware penalties with \emph{CUT-U (p)} and \emph{CUT-S (p)}, correspondingly.

\section{Experiments}

\subsection{Dataset}

We conducted our experiments on two different simplification datasets, the summary statistics of which are presented in Table \ref{tab:datasets}.

\begin{table}[h]
\centering
\begin{adjustbox}{max width=0.49\textwidth}
\begin{tabular}{lccc}
\hline
\textbf{} & \textbf{WikiLarge} & \textbf{Newsela} \\
\hline
Source & 291,402 & 81,705 \\
Target & 291,402 & 76,073 \\
Train & 5,000 & 5,000  \\
Valid & 2,000 & 1,500 \\
Test & 359 & 1,500 & - \\
Vocab (src)  & 41,303 & 33,316 \\
Vocab (tgt) & 39,912 & 22,405  \\
Compression & 0.98 & 0.76 \\
FKGL (src) & 9.51 & 8.51 \\
FKGL (tgt) & 6.33 & 2.86 \\
\hline
\end{tabular}
\end{adjustbox}
\caption{Statistical description of the datasets}
\label{tab:datasets}
\vspace{-0.5cm}
\end{table}


\textbf{WikiLarge} has become a benchmark for training and evaluating text simplification models \cite{zhang-lapata-2017-sentence}. Originally it had 296,402 sentence pairs, but we took 5,000 pairs for machine translation step during our model training. For validations and tests, we used TurkCorpus \cite{xu-etal-2016-optimizing}.

\textbf{Newsela} is a corpus of thousands of news articles professionally leveled to different reading complexities \cite{xu-etal-2015-problems}. We used the most contrast article versions. For the machine translation step, for the test, and for the validation datasets, we used parallel complex-simple pairs provided by \cite{xu-etal-2015-problems}.


\subsection{Metrics}

We use a variety of well established metrics from \emph{Easier Automatic Sentence Simplification Evaluation} (EASSE) framework~\cite{alva-manchego-etal-2019-easse} to analyze the quality of the produced text simplifications, including:  

\textbf{BLEU} (Bilingual Evaluation Understudy) - a precision-oriented metric that estimates the proportion of \textit{n}-gram matches between a system’s output and a reference \cite{papineni-etal-2002-bleu}.

\textbf{SARI}, introduced by \cite{xu-etal-2016-optimizing}, compares system output against the references and against the input sentence. It measures how the simplicity of a sentence was improved based on the words added, deleted, and kept by the system.
    
\textbf{FKGL} (Flesch-Kincaid Grade Level) estimates the readability of text using cognitively motivated features \cite{Kincaid1975DerivationON}. Commonly reported as measures of simplicity, FKGL relies on average sentence lengths and the number of syllables per word.
    
We complement this set with (a) the proportion of \textbf{exact matches} between simplified and original sentences, (b) the average proportion of \textbf{added words} and (c) the average proportion of \textbf{deleted words} to gain additional insights on the nature of the transformations that a model is performing on the input texts.

\begin{table}[h]
\centering
\begin{adjustbox}{max width=0.48\textwidth}
\begin{tabular}{lcccccc}
\hline
 & BLEU & SARI & FKGL & Match & Add & Del \\
\hline
PBMT-R & 18.19 & 15.77 & 7.59 & - & - & - \\
Hybrid & 14.46 & 30.00 & 4.01 & - & - & - \\
EncDecA & 21.7.0 & 24.12 & 5.11 & - & - & - \\
DRESS & 23.21 & 27.37 & 4.13 & - \\
DRESS-LS & \textbf{24.30} & 26.63 & 4.21 & - & - & - \\
DMASS+DCSS  & - & 27.28 & 5.17 & - & - & - \\
\hline
XLM-U & 16.97 & 19.32 & 10.52 & 0.46 & 0.04 & 0.05 \\
CUT-U (p) & 17.21 & 16.51 & 8.8 & 0.27 & 0.03 & 0.04\\
CUT-U (t) & 18.78 & 37.87 & 6.55 & 0.05 & 0.07 & 0.41 \\
\hline
XLM-S & 19.44 & 43.18 & 4.18 & 0.09 & 0.12 & 0.53 \\
CUT-S (p) & 21.33 & 42.47 & \textbf{3.65} & 0.07 & 0.1 & 0.52 \\
CUT-S (t) & 21.70 & \textbf{46.88} & 3.92 & 0.04 & 0.16 & 0.58 \\
\hline
Output = Input & 18.52 & 12.78 & 10.36 & 1.00 & 0.00 & 0.00 \\
Output = Ref & 100.00 & 100.00 & 4.18 & 0.00 & 0.19 & 0.61 \\
\hline
\end{tabular}
\end{adjustbox}
\caption{Performance comparison of CUT models and baselines on Newsela dataset.}
\label{tab:newsela-results}
\vspace{-0.5cm}
\end{table}

\begin{table*}[t]
\centering
\begin{adjustbox}{max width=1\textwidth}
\begin{tabular}{ll}
\hline
Input & Back in 1950 , Eiji Toyoda visited a Ford plant to learn how Americans made cars . \\
Reference & He visited a Ford factory back in 1950 to learn how Americans made cars . \\
\hline
$NbChars_{1.0}+LevSim_{1.0}$ & Back in 1950 , Eiji Toyoda visited a Ford plant to learn how Americans made cars . \\
$NbChars_{1.0}+LevSim_{0.75}$ & In 1950 , Eiji Toyoda visited a Ford factory to learn how Americans made cars . \\   
$NbChars_{1.0}+LevSim_{0.5}$ & In 1950 , Eiji Toyoda visited a Ford factory to learn how to make cars . \\          
$NbChars_{1.0}+LevSim_{0.25}$ & In 1950 , Eiji Toyoda visited a Ford factory . \\
\hline
$NbChars_{1.0}+WordRank_{1.0}$ & In 1950 , Eiji Toyoda visited a Ford factory to learn how Americans made cars . \\   
$NbChars_{1.0}+WordRank_{0.75}$ & In 1950 , Eiji Toyoda visited a Ford factory to learn how to make cars . \\          
\hline
\end{tabular}
\end{adjustbox}
\caption{Results of applying control tokens to an example in a Newsela test set: \emph{NbChars} for compression ratio, \emph{LevSim} for Levenshtein similarity, \emph{WordRank} as defined in \cite{martin2019controllable}.}
\label{tab:newsela-tokens-results}
\vspace{-0.5cm}
\end{table*}

\subsection{Baselines}

We benchmarked our model against several well-known baselines:

\textbf{PBMT-R} is phrase-based machine translation system with a re-ranking post-processing step proposed by \cite{wubben-etal-2012-sentence} 

\textbf{Hybrid} is a simplification model that includes a probabilistic model for splitting and dropping and a PBMT-R model for substitution and reordering \cite{narayan-gardent-2014-hybrid} 

\textbf{SBMT-SARI} is a syntax-based translation model trained on PPDB \cite{ganitkevitch-etal-2013-ppdb} and trained with SARI  \cite{zhang-lapata-2017-sentence}

\textbf{EncDecA}, a basic attention-based encoder-decoder model,  \textbf{DRESS}, a deep reinforcement learning model, {DRESS-LS}, a linear combination of DRESS and the lexical simplification model, all of them were introduced in \cite{zhang-lapata-2017-sentence}. 

\textbf{DMASS+DCSS} is a combination of DMASS and DCSS models from \cite{zhao2018integrating}.

\textbf{UNTS+10K} is an unsupervised model based on a shared encoder and two decoders with limited supervision of 10K labeled examples \cite{surya-etal-2019-unsupervised}.

The BLEU, SARI, and FKGL results for the above-mentioned models were taken from \cite{zhang-lapata-2017-sentence} and \cite{zhao2018integrating}. 

\subsection{Results}


In Table~\ref{tab:newsela-results}, we summarize the results of the experiments on the Newsela dataset. Both control mechanisms -- penalties and tokens -- achieve superior performance in comparison to the unsupervised baseline defined by the off the shelf XML-U model, however, a learning-based approach (\emph{CUT-U (t)}) outpeforms a decoding-based one (\emph{CUT-U (p)}). This finding is in-line with the previous results achieved for supervised text summarization \cite{kikuchi2016controlling}. 

\emph{CUT-U (t)} out-performs all baseline models from the literature (including both supervised and unsupervised) on the SARI metric. Moreover, when we combine our unsupervised control mechanisms with extra supervision provided by the MT step in XLM model, we achieve the highest SARI score of $46.88$ (\emph{CUT-S (t)}) and the lowest FKGL of $3.65$ (\emph{CUT-S (p)}) across all compared methods. 

In Table~\ref{tab:newsela-tokens-results}, we demonstrate an example of applying control tokens to an input sentence in the Newsela test set. A combination of different tokens provides flexibility to adjust the sentence's output complexity, which is in line with the previous results in \cite{martin2019controllable}. What is remarkable, however, is that the model has managed to train itself in a completely unsupervised way by gradually learning from noisy outputs of the back-translation iterations.  

Our results have been less striking on the other popular benchmark -- the WikiLarge dataset Table~\ref{tab:wikilarge-results}. Although our models achieved significantly better results on SARI and FKGL than both \emph{XLM-U} and \emph{XLM-S} baselines, as well as significantly decreased the exact match ratio, we have not observed equivalent improvements in the BLEU score and in comparison to the-state-of-the-art results. We attribute this to the fact that the WikiLarge dataset is a significantly noisier dataset \cite{xu-etal-2015-problems}. We will be looking in understanding and improving our performance in these settings in future work.

\begin{table}[h]
\centering
\begin{adjustbox}{max width=0.48\textwidth}
\begin{tabular}{lcccccc}
\hline
 & BLEU & SARI & FKGL & Match & Add & Del \\
\hline
PBMT-R & 81.11 & 38.56 & 8.33 & - & - & - \\
Hybrid & 48.97 & 31.40 & \textbf{4.56} & - & - & - \\
SBMT-SARI & 73.08 & 39.96 & 7.29 & - & - & - \\
EncDecA & 88.85 & 35.66 & 8.41 & - & - & - \\
DRESS & 77.18 & 37.08 & 6.58 & - & - & - \\
DRESS-LS & 80.12 & 37.27 & 6.62 & - & - & - \\
DMASS+DCSS  & - & \textbf{40.42} & 7.18 & - & - & - \\
UNTS+10K & 76.13 & 35.29 & - & - & - & - \\
\hline
XLM-S & 92.66 & 30.99 & 9.68 & 0.73 & 0.02 & 0.02 \\
CUT-S (t) & 41.33 & 32.01 & 8.73 & 0.05 & 0.16 & 0.19 \\
CUT-S (t+p) & 78.01 & 35.64 & 8.01 & 0.18 & 0.04 & 0.22 \\
\hline
XLM-U & \textbf{94.83} & 28.30 & 9.75 & 0.76 & 0.02 & 0.01 \\
CUT-U (t) & 49.70 & 23.89 & 9.77 & 0.30 & 0.04 & 0.06 \\
\hline
Output = Input & 97.41 & 27.32 & 9.90 & 1.00 & 0.00 & 0.00 \\
Output = Ref & 68.87 & 40.83 & 8.33 & 0.00 & 0.19 & 0.21 \\
\hline
\end{tabular}
\end{adjustbox}
\caption{Performance comparison of CUT models and baselines on Wikipedia Large.}
\label{tab:wikilarge-results}
\vspace{-0.5cm}
\end{table}

\section{Conclusions}

In this paper, we looked at two unsupervised mechanisms -- a learning-based and a decoding-based -- to control the output complexity of text simplification algorithms. We built on an idea of adding complexity control tokens in the input text and applied it to the back-translation algorithm. By iterating the procedure of generating a noisy translation of a sentence and learning from its relative complexity compared to the original, the model self-supervised its ability to produce a controllable output and improved its simplification performance overall. An alternative -- decoding-based mechanism -- has also improved in comparison with the baseline but has demonstrated inferior performance on SARI and BLEU metrics. 

While we find our models' ability to self-supervise on noisy outputs rather striking, we think there is even more potential for improving the performance of this mechanism by providing more guidance in the trial-and-error process of applying control sequences and learning from them with reinforcement learning. We also plan to explore the generalizability of the unsupervised approaches by conducting a cross-corpora validation, i.e., validate the models on the corpora they have not seen during training. We believe that together these will help to establish unsupervised text simplification as a viable alternative to the supervised methods and remove the constraints imposed by the necessity of compiling parallel text corpora. 

\bibliographystyle{acl_natbib}
\bibliography{emnlp2020}

\end{document}